\documentclass{article}

\usepackage{microtype}
\usepackage{graphicx}
\usepackage{subcaption}
\usepackage{booktabs} 

\usepackage{hyperref}

\usepackage[preprint]{icml2026}

\usepackage{amsmath,amsfonts,bm}

\def\eqref#1{equation~\ref{#1}}

\def\1{\bm{1}}

\DeclareMathAlphabet{\mathsfit}{\encodingdefault}{\sfdefault}{m}{sl}
\SetMathAlphabet{\mathsfit}{bold}{\encodingdefault}{\sfdefault}{bx}{n}

\def\ident{{\mathbbm{1}}}

\newcommand{\E}{\mathbb{E}}

\usepackage{hyperref}
\usepackage{graphicx}
\usepackage{enumitem}
\usepackage{booktabs}
\usepackage{multirow}
\usepackage{array}
\usepackage{url}
\usepackage{makecell}
\usepackage{listings}
\usepackage{algorithm}
\usepackage{algpseudocode}
\usepackage{amsmath}
\usepackage{amsthm}
\usepackage{mathtools}
\usepackage{adjustbox}
\usepackage{wrapfig}
\usepackage{bbm}
\usepackage{dsfont}

\usepackage[utf8]{inputenc}
\usepackage[T1]{fontenc}
\usepackage{xcolor}
\usepackage{upquote}
\usepackage{listings}
\usepackage{amsthm}

\usepackage{amsthm}
\usepackage{ amssymb }

\newtheoremstyle{upright}%
  {6pt}   
  {6pt}   
  {\normalfont} 
  {}      
  {\bfseries} 
  {.}     
  {0.5em} 
  {}

\theoremstyle{upright}
\newtheorem{corrolary}{Corollary}
\definecolor{pykeyword}{RGB}{0,92,150}   
\definecolor{pystring}{RGB}{156,102,31}  
\definecolor{pycomment}{RGB}{96,128,96}  
\definecolor{pyidentifier}{RGB}{45,45,45}
\definecolor{pynumber}{RGB}{128,64,0}    
\definecolor{lightgray}{gray}{0.97}

\lstdefinestyle{pywrap}{
  language=Python,
  backgroundcolor=\color{lightgray},
  basicstyle=\ttfamily\small,
  numbers=left,
  numberstyle=\tiny\color{gray},
  numbersep=8pt,
  showstringspaces=false,
  keepspaces=true,
  columns=fullflexible,
  breaklines=true,
  breakatwhitespace=true,
  postbreak=\mbox{\textcolor{gray}{$\hookrightarrow$}\space},
  frame=single,
  framerule=0.3pt,
  rulecolor=\color{gray!50},
  xleftmargin=1em,
  xrightmargin=1em,
  aboveskip=0.8\baselineskip,
  belowskip=0.8\baselineskip,
  captionpos=b,
  keywordstyle=\color{pykeyword}\bfseries,
  stringstyle=\color{pystring},
  commentstyle=\color{pycomment}\itshape,
  identifierstyle=\color{pyidentifier},
  emph={self,async,await}, emphstyle=\color{pynumber}\bfseries
}

\usepackage{tcolorbox}
\tcbuselibrary{listings, skins}
\usepackage[table]{xcolor}

\usepackage{bbm}
\usepackage{dsfont}

\usepackage[capitalize,noabbrev]{cleveref}

\theoremstyle{plain}
\newtheorem{theorem}{Theorem}[section]

\newtheorem{lemma}[theorem]{Lemma}

\theoremstyle{definition}

\newtheorem{assumption}[theorem]{Assumption}
\theoremstyle{remark}

\usepackage[textsize=tiny]{todonotes}

\icmltitlerunning{ARM: Discovering Agentic Reasoning Modules for Generalizable Multi-Agent Systems}

\begin{document}

\twocolumn[
  \icmltitle{ARM: Discovering Agentic Reasoning Modules for Generalizable Multi-Agent Systems}



  \icmlsetsymbol{equal}{*}

  \begin{icmlauthorlist}
    \icmlauthor{Bohan Yao}{yyy,comp}
    \icmlauthor{Shiva Krishna Reddy Malay}{comp}
    \icmlauthor{Vikas Yadav}{comp}
  \end{icmlauthorlist}

  \icmlaffiliation{yyy}{University of Washington}
  \icmlaffiliation{comp}{ServiceNow}

  \icmlcorrespondingauthor{Vikas Yadav}{vikas.yadav@servicenow.com}

  \icmlkeywords{Machine Learning, ICML}

  \vskip 0.3in
]



\printAffiliationsAndNotice{}  

\begin{abstract}

Large Language Model (LLM) powered multi-agent systems have achieved state-of-the-art results on various complex reasoning tasks, leading to a growing interest in automatic methods for designing such systems. However, existing approaches often require re-discovering optimal system architectures for each new task and model, while their performance often only matches that of simple baselines such as Chain-of-Thought (CoT) prompting. This suggests that individual steps within a chain of reasoning progress may be an appropriate target for optimization. We propose the \textbf{A}gentic \textbf{R}easoning \textbf{M}odule (ARM), a generalization of stepwise reasoning (such as in CoT), where each granular reasoning step is performed by an agentic workflow that is automatically discovered and optimized through a reflection-guided tree search. Starting from a simple seed CoT module, ARM is evolved via our proposed structured search algorithm over code space, producing powerful reasoning sub-routines that can be composed recursively or integrated in a learned higher-level meta-policy. Across a diverse range of reasoning tasks and foundation models, ARM-based systems consistently outperform both manually designed and automatically designed MAS baselines, while exhibiting strong cross-task and cross-model generalization without further optimization, demonstrating the potential of ARM to produce performant reasoning systems.
 
\end{abstract}

\section{Introduction}

Large Language Model (LLM) powered multi-agent systems (MAS) have achieved strong results on complex reasoning benchmarks \citep{park2023generative, qian2023communicative, hong2023metagpt}. These systems decompose a problem across multiple LLM-powered agents with distinct roles and coordinate them via a meta-agent or a communication protocol \citep{wu2023autogen, li2023camel}. Motivated by their success, recent work has moved from manually designed MAS to \emph{automatic} methods that synthesize agent roles, interaction patterns, and workflow graphs directly \citep{chen2023agentverse, dong2023self, FlowReasoner2025, zhang2024aflow}. However, many discovered systems are optimized at the \emph{task level} and often need to be re-discovered when the task distribution or the underlying foundation model changes \citep{hu2025automated, zhang2025aflow}. Moreover, multiple studies report that these systems frequently only match, or sometimes underperform, simple and well-tuned baselines \citep{wang2024rethinking, yao2025toolbox}. We observe the same trend in our results (Table~\ref{tab:main_results}), suggesting that optimizing full task-level orchestrations alone may not reliably produce transferable gains.

A complementary perspective is to ask where complex reasoning success actually comes from. Many hard benchmarks require \emph{multi-step progress}, where solving the full problem depends on repeatedly executing reliable reasoning steps, maintaining consistency across intermediate states, and limiting error propagation. This connects naturally to Chain-of-Thought (CoT) prompting, which remains a widely used stepwise reasoning technique \citep{wei2022chain}. CoT improves performance by producing intermediate reasoning steps before the final answer \citep{nye2021show, kojima2022large, yao2023tree}. Related frameworks such as ReAct follow the same principle while interleaving reasoning with actions, including tool usage \citep{yao2022react}. More broadly, modern reasoning systems can be viewed as composing a set of \emph{reasoning steps} arranged in a workflow graph. The graph may be sequential, branching, or iterative, but each node still requires a reliable per-step operator that maps the current state to a next-step output \citep{creswell2022selection, chen2023fireact}.

This motivates our central thesis. If the quality of complex reasoning is governed by the reliability of the per-step operator, then MAS should be designed not only at the task level, but also at the \emph{granular reasoning step level}. We propose the \textbf{A}gentic \textbf{R}easoning \textbf{M}odule (ARM), a generalization of CoT where each granular reasoning step is performed by a small, homogeneous agentic workflow that can be invoked repeatedly wherever a reasoning node appears in a workflow graph. ARM exposes a stable interface. It takes the current reasoning state as input and returns the next-step output, optionally producing intermediate artifacts such as critiques, votes, or verification outcomes. This makes ARM a plug-and-play building block. It can replace the standard ``thought'' step in CoT-style traces, and it can also be integrated into broader agentic workflows with tools, planners, or retrievers, including graphs that branch, merge, or run parallel reasoning paths.

We automatically discover and optimize ARM through a reflection-guided tree search over code space. The search starts from a simple seed CoT module and evolves it into stronger reasoning sub-routines. A meta-agent proposes structured mutations and uses reflection over execution traces to identify failure modes and prioritize targeted improvements \citep{shinn2023reflexion}. Crucially, our optimization is designed to provide credit assignment at the step level. Instead of selecting workflows solely by end-task validation performance, we evaluate candidate modules by their effect on local segments of multi-step traces, which encourages improvements to step reliability and reduces brittle overfitting. The resulting ARM modules can be composed recursively, and they can also be integrated into a learned higher-level meta-policy that orchestrates multiple ARM invocations or parallel ARM traces when beneficial.

By focusing on a general-purpose, reusable step module rather than a task-specific heterogeneous graph, ARM targets a core bottleneck of complex reasoning. Across a diverse range of reasoning tasks and foundation models, ARM-based systems consistently outperform both manually designed and automatically designed MAS baselines, while exhibiting strong cross-task and cross-model generalization without further optimization.

Key contributions of our work are as follows:

\begin{itemize}
    \item We present the Agentic Reasoning Module (ARM), a general MAS at the level of an individual reasoning step. ARM can be invoked repeatedly as the fundamental step operator wherever a reasoning node appears in a workflow graph, including sequential, branching, and parallel topologies.

    \item We propose a reflection-guided tree search over code space that automatically discovers and optimizes ARM starting from a seed CoT module, yielding powerful reasoning sub-routines that can be composed recursively or integrated within a learned higher-level meta-policy.

    \item We demonstrate that ARM-based systems outperform strong manually designed and automatically designed MAS baselines across diverse reasoning tasks and foundation models, while maintaining strong cross-task and cross-model generalization without re-optimization.
\end{itemize}

\section{Related Works}

\paragraph{Single-Agent and Multi-Agent Reasoning Systems}

The landscape of LLM-based reasoning is broadly divided into single-agent and multi-agent paradigms. Single-agent systems have demonstrated remarkable capabilities by augmenting the core LLM with sophisticated reasoning and action frameworks. A prominent example is the ReAct framework, which interleaves reasoning steps with actions, enabling the agent to interact with external tools like search engines to gather information and refine its reasoning process \citep{yao2023react}. Other approaches have focused on enhancing single agents with self-reflection and memory to learn from past mistakes and improve performance iteratively \citep{shinn2023reflexion, madaan2023self}. While these systems are powerful, their development has often focused on narrower tasks, such as tool-based search, retrieval, and question answering, rather than general-purpose complex reasoning.

In parallel, Multi-Agent Systems (MAS) have emerged as a dominant approach for tackling highly complex problems, often outperforming single-agent counterparts \citep{park2023generative, qian2023communicative}. Frameworks like AutoGen \citep{wu2023autogen}, Camel \citep{li2023camel}, and MetaGPT \citep{hong2023metagpt} orchestrate multiple LLM-powered agents, each assigned a specialized role (e.g., programmer, critic, tester). These agents collaborate, debate, and synthesize information to produce solutions for tasks like software development and complex reasoning. A key characteristic of these systems is their heterogeneous nature; each agent is distinct, with a manually engineered role and persona, connected through a predefined and often complex communication topology. In stark contrast, our ARM-based approach constructs a powerful MAS from homogeneous building blocks. The ARM itself is a self-contained, versatile reasoning module that is applied repeatedly, acting as the fundamental unit of thought for all "agents" in the system, thereby simplifying the design while enhancing generalizability.
\vspace{-1em}
\paragraph{The Surprising Efficacy of Simple Reasoning Baselines}
Despite the architectural complexity of many state-of-the-art MAS, a critical and recurring observation is the surprising competitiveness of simple reasoning baselines. Foundational techniques like Chain-of-Thought (CoT) \citep{wei2022chain}, and simple extensions like Self-Consistency (CoT-SC) which samples multiple reasoning chains and takes a majority vote \citep{wang2022self}, often achieve performance on par with, or even superior to, intricate multi-agent frameworks \citep{zhang2025multi}. This phenomenon is particularly pronounced with the advent of increasingly powerful frontier foundation models \citep{ke2025survey}. As these models develop stronger native reasoning abilities, the high-level conceptual guidance provided by a simple CoT prompt is often sufficient to unlock their full potential, rendering the overhead of complex agent orchestration less impactful. This suggests that the primary bottleneck is not necessarily the high-level orchestration strategy but the quality and robustness of the fundamental, step-by-step reasoning process. Our work is directly motivated by this insight, positing that evolving the core reasoning operator—the "thought" in the chain—is a more fruitful direction than designing ever-more-complex superstructures around a static, simple CoT unit.
\vspace{-1em}

\paragraph{Automated Design of Multi-Agent Systems}
Recognizing the significant manual effort required to design effective MAS, recent research has explored automating this process. Approaches like ADAS \citep{hu2025automated}, Aflow \cite{zhang2025aflow}, and Flow-Reasoner \citep{gao2025flowreasoner} aim to automatically discover the optimal agent roles and their interaction topology for a given task domain. However, these techniques suffer from two major drawbacks. First, they are computationally expensive, requiring a costly re-discovery process for each new task domain. Second, the discovered systems are often highly specialized and brittle, tuned specifically for the validation data of a single domain. As our results will later show, with the latest generation of foundation models, these automatically discovered systems can be outperformed by simple CoT baselines. Our work diverges from this paradigm. Instead of discovering a complex, domain-specific agent topology, we focus on discovering a single, domain-agnostic reasoning module (ARM). This ARM acts as a universal, high-quality building block that provides superior performance and generalizability without the need for task-specific rediscovery, offering a more scalable and robust path forward for MAS design.

\vspace{-1em}
\paragraph{LLM based Prompt Optimizers}
Recent research has focused on LLMs as prompt optimizers, leveraging their generative and reasoning capabilities to automatically improve prompts within a fixed workflow \cite{zhou2022large, yang2024large, khattab2024dspy,guo2024evoprompt,novikov2025alphaevolve,fernando2024promptbreeder}.  Notably,  evolutionary approaches coupled with deep reflection over rollouts, such as in GEPA \cite{agrawal2025gepa}, have been shown to offer significant advantages in sample efficiency compared to methods that involve updating model weights via Reinforcement Learning. Within the MAS framework, these approaches can be viewed as optimizing the individual node (LLM role and prompt) without changing the overall interaction topology. Our proposed approach applies this idea to designing the full granular reasoning block (ARM) as a self contained python module.

\section{Methodology: Discovering the Agentic Reasoning Module}

\begin{figure*}[!htbp]
    \centering
    \includegraphics[width=0.8\textwidth]{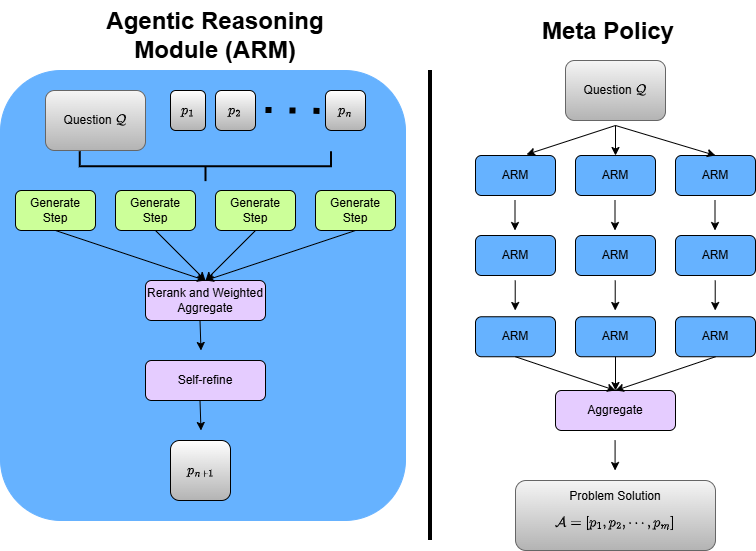}
    \caption{\footnotesize An illustration of the proposed ARM module on the left and the meta policy on the right using "Self refine" as an example MAS. The ARM module takes a question and previous reasoning steps and executes a MAS to get the next step. The meta policy uses ARM as a sub-module and orchestrates the overarching global strategy.  Note that this is for illustration only, the actual step generator and the meta policy discovered by Algorithm \ref{alg:rets} is more complex (See Appendix \ref{apdx:discovered_arms}). }
    \label{fig:ARM_Meta-policyFigure}
    \vspace{-3mm}
\end{figure*}

We introduce the \textbf{Agentic Reasoning Module (ARM)}, a self-contained, code-based multi agentic system designed to execute a single, granular step within a complex reasoning process. ARM is conceived as a structured, agentic replacement for a single step in a Chain of Thought (CoT) sequence \cite{wei2022chain}. While standard CoT prompts an LLM to generate the next reasoning step via naive, monolithic textual generation, an ARM employs an internal multi-agent system (MAS) to produce reasoning steps with greater structure and control. 

Following prior work, \cite{hao2023reasoning}, we define the multi-agentic system as a programming module - a self contained Python function block, while allowing for structured control flow and access to essential APIs such as calling an external LLM, structuring the role and the prompt, and input/output format expectations. Functionally, an ARM accepts the initial problem statement and prior reasoning steps as input, and continues the reasoning until the next logical step in the solution.
\vspace{-0.5em}
\subsection{A Decomposable Framework for Agentic Reasoning}
\vspace{-0.5em}

Let the distribution over problem-solution pairs be $\mathcal{D}$ over $(\mathcal{Q}, \mathcal{A})$. A solution $\mathcal{A}$ consists of a sequence of reasoning steps $[p_1, p_2, \dots, p_N]$, where each step $p_i$ belongs to the space of all possible reasoning steps $\mathcal{P}$. We model the problem-solving process with two key functions:

\textbf{The Step-Generator Module ($m \in \mathcal{M}$):} This is a program that performs a single step of reasoning. It takes the problem question $q \in \mathcal{Q}$ and the history of previous reasoning steps $p_{in} \in \mathcal{P}^*$ as input and returns the next reasoning step $p_{out} \in \mathcal{P}$. Its signature is $m: \mathcal{Q} \times \mathcal{P}^* \to \mathcal{P}$. An \textbf{Agentic Reasoning Module (ARM)} is a structured, code-based implementation of such a module, which can itself be a self-contained MAS.

\textbf{The Meta-Policy ($\pi \in \Pi$):} This is a higher-order program that defines the overarching strategy. It takes a question $q$ and a specific step-generator module $m$ and orchestrates calls to $m$ to generate a complete solution $a \in \mathcal{A}$. Its signature is $\pi: \mathcal{Q} \times \mathcal{M} \to \mathcal{A}$.

Within this framework, standard Chain of Thought (CoT) can be seen as a simple baseline pairing. It uses a basic step-generator, $m_{CoT}$, which is a single call to an LLM, and a simple \textbf{recursive meta-policy}, $\pi_{Rec}$, which applies $m_{CoT}$ repeatedly until a final answer is produced. Our approach independently discovers a more powerful module $m^*$ (the ARM) and a more sophisticated meta-policy $\pi^*$.
\vspace{-0.5em}
\subsection{Discovering the Optimal Step-Generator (\texorpdfstring{$m^*$}{m*})}
\label{method_step_gen}
\vspace{-0.5em}
Our primary goal is to find a step-generator module $m^*$ that is a general-purpose and superior replacement for the simple text generation step in $m_{CoT}$.

We can formalize a single reasoning step as an update function, $U_{m,q}$, that appends the output of module $m$ to the current reasoning history $h$:
$$U_{m,q}(h) = h \cdot [m(q,h)]$$
where $\cdot$ denotes list concatenation. A full, $n$-step reasoning trace generated by the recursive policy $\pi_{Rec}$ is thus the $n$-fold composition of this update function: $\pi_{Rec}(q, m) = U_{m,q}^n(\emptyset)$.

Ideally, we would discover the optimal module $m^*$ by maximizing the expected reward $\mathcal{R}$ over the entire problem-solving trace:
$$m^* = \underset{m\in \mathcal{M}}{\text{argmax}} \quad \mathbb{E}_{(q,a)\sim \mathcal{D}}\left[\mathcal{R}\left(\pi_{Rec}\left(q, m\right), a\right)\right]$$
However, optimizing this objective directly is intractable due to two main challenges:
1.  \textbf{Difficult Credit Assignment:} The reward is observed only at the end of a long sequence of steps, making it difficult to determine which specific application of $m$ was responsible for the final outcome.
2.  \textbf{Unconstrained Search Space:} The space of possible code-based modules $\mathcal{M}$ is vast, making an unguided search highly inefficient.

To address this, we introduce a practical \textbf{scaffolded surrogate objective}. Instead of evaluating $m$ on a full rollout generated by itself, we evaluate it within the stable context of a reference trace generated by the baseline $m_{CoT}$. Specifically, we replace a small, contiguous block of $l$ steps within an $n$-step CoT trace with our candidate module $m$. The optimization problem becomes:
\scriptsize$$m^* = \underset{m\in \mathcal{M}}{\text{argmax}} \quad \mathbb{E}_{(q,a)\sim \mathcal{D}}\left[\mathcal{R}\left(U_{m_{CoT},q}^{n-l-i} \circ U_{m,q}^l \circ U_{m_{CoT},q}^{i}(\emptyset), a\right)\right]$$\normalsize
where $n = |\pi_{Rec}(q, m_{CoT})|$ is the length of the reference CoT trace, and the starting index $i$ is chosen randomly from $[0, n-1]$. This formulation isolates the performance contribution of $m$ to a small window, enabling direct credit assignment. Furthermore, the surrounding CoT context provides a powerful inductive bias, constraining the search to modules that behave as effective, incremental reasoning steps. This mirrors the conservative policy-improvement principle \cite{10.5555/645531.656005}, where a candidate policy is evaluated under a stable reference distribution to guarantee monotonic improvement. Likewise, the scaffold constrains module updates within a fixed Chain-of-Thought context, ensuring stable, incremental reasoning gains. We pick $l=3$ as it is shortest window which is  sufficient to expose the module $m$ to critical compositional patterns—$(U_{m_{CoT}, q} \circ U_{m,q})$, $(U_{m,q} \circ U_{m,q})$, and $(U_{m,q} \circ U_{m_{CoT},q})$—while keeping the optimization tractable.
\vspace{-0.5em}
\subsection{Discovering the Optimal Meta-Policy (\texorpdfstring{$\pi^*$}{π*})}
\label{method_meta_policy}
\vspace{-0.5em}
While an optimized step-generator $m^*$ improves the quality of each reasoning step, the high-level strategy $\pi$ that orchestrates these steps is equally critical. A simple recursive policy, $\pi_{Rec}$, may be suboptimal for complex problems that could benefit from strategies like parallel rollouts (for self-consistency)  or iterative refinement loops \cite{wang2022selfconsistency, madaan2023self}.

Searching for an optimal meta-policy $\pi^*$ by repeatedly evaluating candidates with the full, complex $m^*$ module is computationally prohibitive. Therefore, we adopt a surrogate-based approach here as well. We search for the optimal meta-policy $\pi^*$ using the fast and computationally cheap baseline step-generator, $m_{CoT}$, as a stand-in for $m^*$.

This zero-shot transfer from $m_{CoT}$ to $m^*$ is effective because our step-generator optimization process (Section 3.2) is explicitly designed to produce an $m^*$ that functions as a superior, "drop-in" replacement for $m_{CoT}$.  A meta-policy that effectively orchestrates the simple steps of $m_{CoT}$ is thus highly likely to generalize to orchestrating the more powerful, but functionally analogous, steps of $m^*$. This allows us to efficiently explore the space of strategies, discovering sophisticated control flows like branching for parallel thought generation or conditional loops for verification, without incurring the high computational cost of using $m^*$.
\vspace{-0.5em}
\subsection{Reflection-Guided Evolutionary Search}
\vspace{-0.5em}
\label{evol_search}
We discover both the optimal step-generator $m^*$ and meta-policy $\pi^*$ using a unified \textbf{Reflection-Guided Evolutionary Search} algorithm. This algorithm performs a tree search over the programmatic space of valid Python modules, where each node in the tree represents a specific program. The search begins with a root node representing the baseline program ($m_{CoT}$ for the step-generator search and $\pi_{Rec}$ for the meta-policy search). The search then iteratively performs three steps:

1.  \textbf{Selection:} A parent node (program) $p_{parent}$ is sampled from the current tree $\mathcal{T}$ using temperature sampling based on it's validation performance.
\vspace{-0.1em}

2.  \textbf{Expansion:} A new child program is generated by a \textbf{Reviewer Agent}, an LLM-based agent that reflects on the parent program's execution traces, correctness, and mutation history to propose a targeted code modification.
\vspace{-0.1em}

3.  \textbf{Evaluation:} The newly generated program is evaluated to obtain its average reward $\bar{\mathcal{R}}$. For a step-generator module, we use the scaffolded objective from Section 3.2. For a meta-policy, we evaluate its performance on a full problem rollout using $m_{CoT}$ as the step-generator.

This entire process is summarized in Algorithm \ref{alg:rets}.

\subsubsection{The Reviewer Agent}

The expansion step is driven by a two-stage \textbf{Reviewer Agent} that intelligently mutates existing programs. This agent consists of two LLM-based components:

\textbf{Critic:} The Critic analyzes execution traces from the parent program. It identifies logical errors, inefficiencies, or patterns of failure, providing a concise, natural-language analysis of the program's strengths and weaknesses.

\textbf{Designer:} The Designer acts as the mutation operator. It takes the original program's code, its performance history, and the Critic's analysis as input. Based on this information, it proposes a single, targeted code modification aimed at addressing the identified issues, generating a complete, syntactically valid Python class for the new program.

This reflection-driven process ensures that the search evolves programs purposefully, rather than through random mutations, leading to more efficient discovery of high-performance modules and policies. The prompts used for the Critic and Designer are detailed in the Appendix.

\section{ARM Search Algorithm}
Algorithm \ref{alg:rets} provides the full pseudocode of the reflection-guided search algorithm for evolving ARM modules.
\vspace{-1em}
\section{Experiments}

\subsection{Benchmarks}

We evaluated our baselines and approach on multiple complex reasoning datasets. To assess complex mathematical reasoning capabilities, we utilized widely studied \textit{American Invitational Mathematics Examination} (\textit{AIME}\footnote{ \url{https://huggingface.co/datasets/MathArena/aime_2025}}) and the \textit{Harvard-MIT Mathematics Tournament} (\textit{HMMT}\footnote{ \url{https://huggingface.co/datasets/MathArena/hmmt_feb_2025}}) datasets. For reasoning evaluations on specialized scientific knowledge, we used \textit{GPQA}, a benchmark containing graduate-level questions in physics, chemistry, and biology designed to be challenging even for human experts~\citep{rein2023gpqa}. Finally, to measure practical, up-to-date reasoning and robustness against data contamination, we used \textit{LiveBench Reasoning \cite{white2024livebench}} \footnote{ \url{https://huggingface.co/datasets/livebench/reasoning}}, a dynamic benchmark with continuously evolving  questions~\citep{jain2024livecodebench}. To reduce the variance of results due to small dataset sizes, we evaluate pass@8 accuracy on AIME and HMMT.

\subsection{Baselines}
We compare our methodology against two distinct groups of multi-agent systems (MAS) baselines: popular handcrafted MAS systems and leading automated MAS generation approaches.

\subsubsection{Handcrafted Multi-Agent Systems:}

We compare against several strong reasoning baselines. \textbf{Chain of Thought (CoT)} \cite{wei2022chain} serves as the fundamental baseline, solving tasks through iterative textual reasoning. \textbf{CoT-Self Consistency (CoT-SC)} \cite{wang2022selfconsistency} improves upon CoT by generating $n=12$ parallel reasoning rollouts and selecting the final answer via a majority vote. \textbf{Self-Refine} \cite{madaan2023self} employs a feedback loop where a Large Language Model (LLM) iteratively critiques and refines its own output. Lastly, \textbf{LLM-Debate} \cite{du2023improvingfactualityreasoninglanguage} initializes multiple LLM agents with diverse roles to generate different reasoning paths, fostering a debate to converge on a final solution.

\subsubsection{Automated Multi-Agent Systems:} 

These baselines include the two leading code based MAS generation approaches: 
\textbf{ADAS} \cite{hu2025automated} and \textbf{AFlow} \cite{zhang2025aflow}. These methods employ search algorithms to automatically discover the optimal agent roles and their complex interaction topology for a given task domain


We evaluate the performance of ADAS and AFlow using both the original optimization configuration of using a 20\% split of the test dataset as the validation dataset (resulting in a benchmark-optimized MAS for each benchmark) and using the ARM optimization configuration of using the 1000-sample subset of Open-R1-Mixture-of-Thoughts \cite{openr1} as the validation dataset (resulting in a single MAS which we evaluate across all benchmarks without benchmark-specific re-optimization). We denote baselines of the former configuration using \emph{``(test set)''} and baselines of the latter configuration using \emph{``(1000-sample)''} in the main results in Table~\ref{tab:main_results}.

\subsection{Models}
We use OpenAI's o4-mini-high \cite{openai2025o3o4mini} reasoning model as the MAS designer for both the baselines ADAS, AFlow, and our method ARM, as MAS generation requires frontier performance in coding, and instruction following. During validation and inference, we three models as backbone LLMs executing the MAS: two closed source models GPT-4.1-nano \cite{openai2025gpt41}, GPT-4o \cite{openai2024gpt4ocard} and one open source model Llama-3.3-70B \cite{meta2024llama3p3}. 

\subsection{Training}
Our training process is designed to independently discover the two core components of our framework: the optimal step-generator module ($m^*$) and the optimal meta-policy ($\pi^*$). This decoupled approach allows us to first forge a powerful, general-purpose reasoning module and then learn a sophisticated strategy to orchestrate it, all without requiring expensive, domain-specific annotations.

\textbf{Validation Dataset: } For both discovery processes, we utilize the a subset (1000 samples) of the Math and Science splits of the Open-or-Mixture-of-Thoughts \cite{openr1} dataset, a general-purpose instruction-following dataset. Our method requires only a one-time, domain-agnostic training phase. The same resulting code artifacts are then deployed across all benchmark domains and foundation models without any task-specific fine-tuning or re-optimization, underscoring the robustness and versatility of our method.

\textbf{Step-Generator ($m^*$) Discovery:} We discover the ARM module by employing the Reflection-Guided Evolutionary Search detailed in Algorithm 1. The search is initialized with a basic Chain-of-Thought module ($m_{CoT}$) and iteratively evolves it by maximizing the scaffolded surrogate objective from Section 3.2. This objective evaluates candidate modules within the context of a baseline CoT trace, enabling efficient and stable optimization.

\textbf{Meta-Policy ($\pi^*$) Discovery:} The meta-policy is discovered independently using the same evolutionary search algorithm. To ensure computational tractability, this search is performed using the simple and fast baseline module, $m_{CoT}$, as a surrogate for the more complex $m^*$ (as justified in Section 3.3). This allows us to efficiently explore the space of high-level strategies and discover a sophisticated meta-policy that can be seamlessly paired with the optimized ARM module.

\section{Results}

\begin{table*}[!ht]
\centering
\resizebox{0.9\textwidth}{!}{%
\begin{tabular}{>{\raggedright\arraybackslash}m{1.1cm} l lc c c c c}
\toprule
\textbf{Model} & \textbf{Method} &  \textbf{MATH-500}&\textbf{AIME25}& \textbf{HMMT25}& \textbf{GPQA} & \textbf{LiveBench} & \textbf{Average} \\
\midrule
\multirow{8}{*}{\rotatebox[origin=c]{90}{GPT-4.1-nano}} 
& CoT &  82.0\%&15.1\% & 9.9\% & 50.0\% & 33.1\% & 38.0\%\\
& CoT-SC &  \textbf{86.2\%}&\underline{21.9\%} & 13.5\% & 50.6\% & 36.9\% & 41.8\%\\
& Self-Refine &  84.2\%&17.2\% & 9.4\% & 50.0\% & 28.1\% & 37.8\%\\
& LLM-Debate &  84.2\%&15.1\% & \underline{16.7\%} & 52.5\% & 33.8\% & 40.5\%\\
\cmidrule(lr){2-8}& ADAS (test set)&  79.8\%&12.0\% & 5.2\% & 48.1\% & 31.2\% & 35.3\%\\
 & ADAS (1000-sample)&  77.3\%&0.0\%& 6.8\%& 46.8\%& 29.4\%&32.0\%\\
& AFlow (test set)&  74.5\%&18.8\% & 12.0\% & 39.9\% & 30.6\% & 35.2\%\\
 & AFlow (1000-sample)&  77.0\%&16.7\%& 10.4\%& 51.3\%& 30.6\%&37.2\%\\
\cmidrule(lr){2-8} & \cellcolor{green!10} ARM \textbf{(Ours)}  & \cellcolor{green!10} 82.0\%&\cellcolor{green!10}18.2\% &\cellcolor{green!10} 14.6\% &\cellcolor{green!10} \underline{60.1\%} &\cellcolor{green!10} \underline{39.4\%}  &\cellcolor{green!10} \underline{42.9\%}\\
& \cellcolor{green!10}ARM + MP \textbf{(Ours)} & \cellcolor{green!10} \underline{86.0\%}&\cellcolor{green!10}\textbf{23.4\%} & \cellcolor{green!10}\textbf{22.4\%} & \cellcolor{green!10}\textbf{61.4\%} & \cellcolor{green!10}\textbf{45.6\%} & \cellcolor{green!10}\textbf{47.8\%}\\ 
\midrule
\multirow{8}{*}{\rotatebox[origin=c]{90}{GPT-4o}}
& CoT &  75.0\%&7.3\% & 0.5\% & 53.8\% & 46.2\% & 36.6\%\\
& CoT-SC &  \underline{81.8}\%&12.5\% & 2.1\% & 53.2\% & 42.5\% & 38.4\%\\
& Self-Refine &  77.2&6.8\% & 2.6\% & 53.8\% & 37.5\% & 35.6\%\\
& LLM-Debate &  \underline{81.8}\%&9.9\% & 3.1\% & 56.3\% & \underline{47.5\%} & 39.7\%\\
\cmidrule(lr){2-8}& ADAS (test set)&  65.5\%&1.0\% & 0.0\% & 46.2\% & 38.8\% & 30.3\%\\
 & ADAS (1000-sample)&  69.0\%&0.0\%& 0.5\%& 46.8\%& 41.9\%&31.6\%\\
& AFlow (test set)&  75.5\%&9.9\% & 3.6\% & 53.8\% & 41.9\% & 36.9\%\\
 & AFlow (1000-sample)&  48.8\%&9.4\%& 0.0\%& 50.6\%& 45.0\%&30.8\%\\
\cmidrule(lr){2-8}&\cellcolor{green!10} ARM \textbf{(Ours)}  &\cellcolor{green!10}  78.3\%&\cellcolor{green!10}\underline{13.5\%} & \cellcolor{green!10}\underline{5.7\%} & \cellcolor{green!10}\underline{59.5\%} & \cellcolor{green!10}\underline{47.5\%} & \cellcolor{green!10}\underline{40.9\%}\\
&\cellcolor{green!10} ARM + MP \textbf{(Ours)} & \cellcolor{green!10} \textbf{82.0\%}&\cellcolor{green!10}\textbf{17.2\%}& \cellcolor{green!10}\textbf{9.4\%}& \cellcolor{green!10}\textbf{60.1\%} & \cellcolor{green!10}\textbf{51.9\%} & \cellcolor{green!10}\textbf{44.1\%}\\ 
\midrule
\multirow{8}{*}{\rotatebox[origin=c]{90}{LLaMA-3.3-70B}}
& CoT &  75.0\%&6.8\% & 3.1\% & 50.0\% & 38.1\% & 34.6\%\\
& CoT-SC &  78.5\%&4.2\% & \underline{5.7}\% & \textbf{53.2\%}& 45.0\% & 37.3\%\\
& Self-Refine &  77.8\%&6.8\% & 4.2\% & \underline{51.3\%} & \underline{46.9\%} & 37.4\%\\
& LLM-Debate &  79.0\%&5.7\% & 4.2\% & 50.6\% & 46.2\% & 37.1\%\\
\cmidrule(lr){2-8}& ADAS (test set)&  67.2\%&3.1\%& 0.0\%& 47.5\%& 37.5\%& 31.0\%\\
 & ADAS (1000-sample)&  22.2\%&3.1\%& 0.5\%& 42.4\%& 46.2\%&22.9\%\\
& AFlow (test set)&  65.2\%&4.7\%& 0.0\%& 46.8\%& 38.1\% & 31.0\%\\
 & AFlow (1000-sample)&  63.2\%&7.2\%& 3.1\%& 46.8\%& 15.6\%&27.2\%\\
\cmidrule(lr){2-8}&\cellcolor{green!10} ARM \textbf{(Ours)}  &  \cellcolor{green!10}\underline{80.0}\%&\cellcolor{green!10}\textbf{8.3\%}&\cellcolor{green!10} 5.2\%&\cellcolor{green!10} 49.6\%&\cellcolor{green!10} 46.2\%& \cellcolor{green!10}\underline{37.9\%}\\
&\cellcolor{green!10} ARM + MP \textbf{(Ours)} &\cellcolor{green!10}  \textbf{80.8\%}&\cellcolor{green!10}\underline{7.8\%}&\cellcolor{green!10} \textbf{6.8\%}& \cellcolor{green!10}50.0\%&\cellcolor{green!10} \textbf{50.0\%}&\cellcolor{green!10} \textbf{39.1\%}\\ 
\bottomrule
\end{tabular}
}
\vspace{2mm}
\caption{\footnotesize Main results on four complex reasoning benchmarks across three foundation models. We compare against two groups of baselines: (1) foundational reasoning strategies used to build agentic systems (CoT, CoT-SC, Self-Refine, and LLM-Debate), and (2) existing state-of-the-art automatic MAS design methods (ADAS and AFlow). Our approach is presented in two variants: \textbf{ARM}, which recursively applies the discovered reasoning module, and our full method, \textbf{ARM + MP}, which combines the ARM with a learned Meta-Policy (MP). Best score in each category is \textbf{bolded} and second best score is \underline{underlined}.}
\label{tab:main_results}
\vspace{-6mm}
\end{table*}

We summarize our results in Table \ref{tab:main_results} and the key findings are as follows:

\begin{enumerate}[label={\bf(\arabic*)},itemsep=0em,topsep=0em,  wide, labelindent=0pt]
\item {\bf Naive Operators outperform MAS:} Simple basic operators such as CoT, Self-refine, LLM-Debate outperform complex MAS systems like AFlow and ADAS. This highlights an important concern regarding the practicality of recent advancemenets in MAS. On the other hand, simple reasoning operators such as CoT perform substantially better across tasks, and varied families of LLMs. Our ARM based reasoning approach is step forward to revitalize traditional yet strong reasonig methods like CoT, by advancing their reasoning steps with agentic blocks. Our ARM based approach further improves upon the CoT performance and achieves best results all the datasets. 

\item {\bf ARM achieving top performance:} ARM consistently outperforms all of the operator baselines. Specifically, in complex datasets such as AIME and HMMT, ARM consistently outperforms existing MAS approaches and all the existing baseline operators. This emphasizes the benefits and strong potential of revitalizing proven traditional reasoning methods like CoT.  

\item {\bf Effects from stronger foundation LLM:} We first note an important observation that with stronger LLMs such as GPT-4o, simple operators such as CoT and CoT-SC outperform complex MASes. Our ARM based reasoning approach further pushes the best performance over the baseline operators with both recent stronger frontier models such as GPT4.1-nano / GPT-4o and older benchmark models such as LLaMa-3.3-70B. 

The implementation of the best performing discovered ARM is presented in Appendix \ref{apdx:discovered_arms}, and the implementation of the best performing discovered Meta-Policy is presented in Appendix \ref{apdx:discovered_metapolicies}.

\end{enumerate}
\vspace{-1em}
\section{Analyses}

To understand the sources of ARM's effectiveness, we performed two key analyses. First, we provide empirical evidence that our search objective discovers fundamentally more reliable reasoning modules by minimizing their per-step error rate. Secondly, we show the validity of our efficient, decoupled training strategy by demonstrating that the learned meta-policy transfers zero-shot from a simple surrogate to the final ARM, yielding significant performance gains.
\vspace{-0.75em}
\subsection{Empirical Validation of the Step-Generator Objective}
\label{sec:ablation_error}
\vspace{-0.75em}
To empirically validate our theoretical claim (Appendix~A) that the scaffolded objective minimizes per-step error, we conducted a targeted ablation study. We executed the top five discovered step-generator modules for a single step, starting from \textit{critical reasoning junctures} identified by an LLM-judge (GPT-OSS-20B) within baseline $m_{CoT}$ traces.  The error rate of each single-step output was then evaluated. As shown in Figure~\ref{fig:ARM_Meta-policyFigure}, a module's rank, determined by our objective, strongly correlates with a lower per-step error rate at these critical points. This result confirms that our search process successfully discovers modules that are fundamentally more robust at a granular level, validating the core mechanism behind ARM's performance.

\begin{figure}[htbp]
    \centering
    \begin{minipage}[c]{0.48\textwidth}  
        \centering
        \resizebox{\linewidth}{!}{%
        \begin{tabular}{cccc}\toprule
            \textbf{Meta Policy Name (abbreviated)} & \textbf{CoT Baseline} & \textbf{CoT$\to$Meta} & \textbf{Meta Policy} \\\midrule
            VWASCCoT & 35.1\% & 33.7\% & 42.0\% \\
            CWDCWACCCoT & 37.2\% & 39.3\% & 41.8\% \\
            RVDCCWASCCoT & 33.7\% & 40.0\% & 41.8\% \\
            DRWASCCoT & 35.5\% & 34.9\% & 41.8\% \\
            MBECDCCWASCCoT & 36.3\% & 39.2\% & 41.4\% \\ \bottomrule
        \end{tabular}
        }
        \caption{\small Validation of the meta-policy transfer for top discovered policies. The table compares performance using the simple surrogate $m_{CoT}$ (\textbf{CoT Baseline}) versus the powerful ARM module $m^*$ (\textbf{Meta Policy}). The intermediate \textbf{CoT$\to$Meta} column isolates the performance gain from the superior $m^*$ module alone by evaluating it on states generated by the baseline.}
        \label{tab:meta_analysis}
    \end{minipage}%
    \hfill
    \begin{minipage}[c]{0.5\textwidth}  
        \centering
        \includegraphics[height=5.5cm]{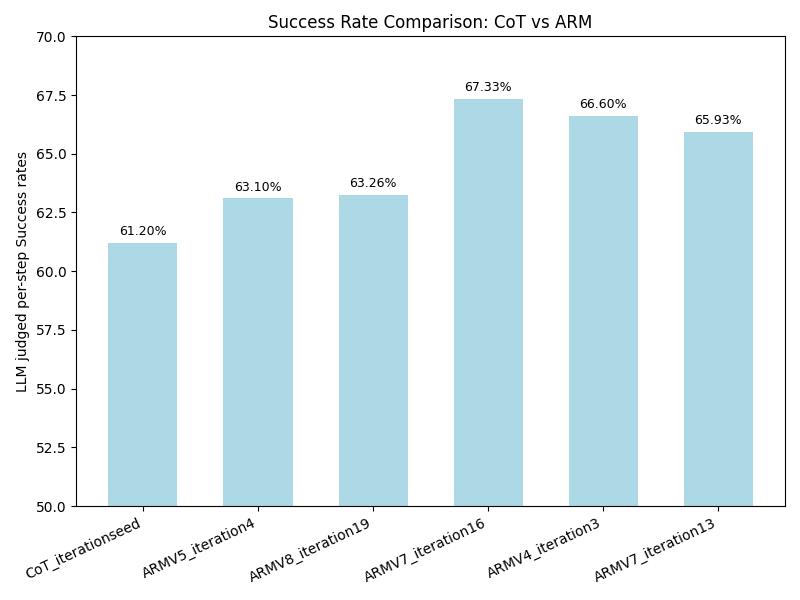}  
        \caption{\small Comparison of LLM judged per-step success rates between the baseline \textit{Chain-of-Thought} (CoT) and multiple \textit{ARM} (CriticChainOfThought) variants. CoT appears first, followed by ARM variants ordered by final performance.}
        \label{fig:arm_analyses}
    \end{minipage}
\end{figure}

\subsection{Empirical Validation of Meta-Policy Transfer}
\label{sec:ablation_metapolicy}
Our methodology relies on a crucial transfer: a meta-policy trained with the simple $m_{CoT}$ module is deployed zero-shot with the powerful, discovered $m^*$ module. The theoretical justification in Appendix~B posits this transfer is effective due to two factors: (1) the inherent superiority of the $m^*$ module, and (2) its ability to guide the reasoning process into more productive states. We designed an experiment to empirically disentangle and verify these two sources of gain.

To do this, we measure and compare three distinct performance configurations. First, we establish a \textbf{baseline performance} using the meta-policy with the simple $m_{CoT}$ module. Second, to isolate the pure \textbf{module improvement gain}, we measure the performance of the powerful $m^*$ module when it takes over from intermediate reasoning states generated by the baseline $m_{CoT}$. Finally, we measure the \textbf{full system performance} of the meta-policy paired with $m^*$ from the start.

The results, shown in Figure~\ref{tab:meta_analysis}, confirm our hypothesis with a clear performance hierarchy. The baseline system performs worst, followed by a significant improvement from simply swapping to the $m^*$ module. The best performance is achieved by the full system, which benefits from both the better module and its ability to find a better reasoning path. This empirically validates the two conditions for successful transfer outlined in Appendix \ref{sec:apdx_theoritical_analysis} and confirms the effectiveness of our decoupled discovery strategy.



\section{Conclusion}

We introduced ARM, a modular agentic reasoning framework that revitalizes the traditional Chain-of-Thought (CoT) paradigm by augmenting it with lightweight agentic blocks. Through extensive experiments, we demonstrated that simple operators such as CoT and Self-Refine not only remain highly competitive but, in many cases, outperform complex Multi-Agent Systems (MAS), highlighting the growing gap between empirical performance and the perceived promise of increasingly elaborate MAS designs. Our results show that ARM consistently advances the performance of CoT across diverse reasoning tasks and model families, establishing top-performing results.  


Beyond empirical improvements, ARM sheds light on an important perspective: improving the granular step by step reasoning process holds the key to progress in reasoning systems. By preserving the simplicity and generality of CoT steps, while enhancing its reasoning depth and modularity, ARM provides a versatile and powerful foundation that can be applied across tasks and models. ARM represents a step toward a robust and broadly applicable modular reasoning approach with LLMs, paving the way for future research to focus on discovering powerful, reusable reasoning units as a core component of agentic systems.


\bibliography{example_paper}
\bibliographystyle{icml2026}

\newpage
\appendix
\onecolumn

\section{ARM Search Algorithm}
Algorithm \ref{alg:rets} provides the full pseudocode of the reflection-guided search algorithm for evolving ARM modules.

\begin{algorithm}[h!]
\caption{Reflection-Guided Search}
\label{alg:rets}
\begin{algorithmic}[1]
\State \textbf{Input:} Initial program $p_{root}$ (e.g., $m_{CoT}$ or $\pi_{Rec}$), evaluation function $\textsc{Evaluate}(\cdot)$, total iterations $K$, exploration constant $C$.
\State \textbf{Initialize:}
\State Tree $\mathcal{T}$ with a single node for $p_{root}$.
\State $p_{root}.\bar{\mathcal{R}} \leftarrow \textsc{Evaluate}(p_{root})$ \Comment{Evaluate the baseline program on a validation batch}
\State $p_{root}.N \leftarrow 1$ \Comment{Initialize visit count for the root}

\For{$t=1$ \textbf{to} $K$}
    \State \Comment{\textit{1. Select a parent program to mutate}}
    \State  $P(p_i)\leftarrow \frac{\exp\left( p_i.\bar{\mathcal{R}} / T \right)}{\sum_{j \in \mathcal{T}} \exp\left( p_j.\bar{\mathcal{R}} / T \right)}$
    \State $p_{parent} \leftarrow \text{Sample}(\mathcal{T}, P)$
    \State \Comment{\textit{2. Expand the tree via reflection}}
    \State traces $\leftarrow$ \textsc{Execute}($p_{parent}$) \Comment{Collect execution traces}
    \State history $\leftarrow$ \textsc{GetMutationHistory}($p_{parent}$)
    \State $p_{new} \leftarrow \textsc{ReviewerAgent}(p_{parent}, \text{traces}, \text{history})$
    
    \State \Comment{\textit{3. Evaluate the new program}}
    \State $p_{new}.\bar{\mathcal{R}} \leftarrow \textsc{Evaluate}(p_{new})$
    \State $p_{new}.N \leftarrow 1$
    
    \State \Comment{\textit{4. Update tree and statistics}}
    \State $\mathcal{T}.\textsc{AddChild}(p_{parent}, p_{new})$
    \State $p_{parent}.N \leftarrow p_{parent}.N + 1$
\EndFor
\State
\State \textbf{return} $\underset{p_i \in \mathcal{T}}{\text{argmax}} \ (p_i.\bar{\mathcal{R}})$ \Comment{Return the program with the highest empirical reward}
\end{algorithmic}
\end{algorithm}

\section{Theoretical Analysis}
\label{sec:apdx_theoritical_analysis}
A complete theoretical analysis of the multi-agentic system ARM powered by LLMs is intractable due to the complex, high-dimensional nature of language generation and the non-stationary of the generation process.  Recent research \citep{chang2025rlstartheoreticalanalysisreinforcement,kim2025metastabledynamicschainofthoughtreasoning} models sequential CoT reasoning steps as a Markov Decision Process by abstracting away the underlying complexities of the text generation process and focusing on higher level reasoning states. Therefore, to build a formal intuition for the design choices in our scaffolded search for the step-generator, and the decoupled search for the meta-policy (Algorithm\ref{alg:rets}), we also analyze an idealized formulation of the problem as a Markov Decision Process (MDP).

Our analysis is particularly inspired by recent work on self taught reasoners (RL-Star) by \cite{chang2025rlstartheoreticalanalysisreinforcement}, where they introduce a step indexed competence parameter $\delta_{t,n}$ which quantifies the advantage in probability of a correct reasoning step at step $n$ during training iteration $t$ over a baseline random reasoner. They show the conditions under which a bootstrapped RL learning algorithm based on rejection sampling shows monotonic improvement and convergence. While our goals are similar (improving the reasoning process), our problem statement has critical differences which makes a straight forward adaption infeasible: RL-Star analyses a system where the LLM's parametric weights are updated via reinforcement learning. On the other hand, ARM treats the LLM as a black box and performs discrete, evolutionary search \citep{fernando2024promptbreeder,agrawal2025gepa} over programming modules that orchestrate calls to the LLM. Consequently, our search is inherently \emph{discrete}, so smoothness-based guarantees do not apply. Hence we do not assume or prove convergence guarantees, and instead motivate the intuition of our scaffolded search process as a conservative policy improvement (CPI) \citep{10.5555/645531.656005} that preferentially selects modules with higher competence leading to improved reasoning process.

\subsection{An Idealized MDP Model of Step-wise Reasoning}
We model the reasoning process as a Markov decision process (MDP) \cite{10.5555/3312046} $\mathcal{M} = (S, N, A, P, R, \gamma)$:
\begin{itemize}
    \item \textbf{State Space $(S)$:} The state space $S=\mathcal{U}\cup \mathcal{G}\cup \mathcal{F}$ is partitioned into three disjoint subsets: 
    \begin{itemize}
        \item $\mathcal{U}$: A state $s\in \mathcal{U}$ represents a partial reasoning trace ${q, p_1, ...p_k}$ that is not yet terminated. 
        \item $\mathcal{G}$: A state $s\in \mathcal{G}$ represents a reasoning path that has successfully ended on the right answer.  In our setting this is when the module emits the \texttt{/boxed\{correct answer\}}. This is an absorbing region. 
        \item $\mathcal{F}$: A state $s\in\mathcal{F}$ represents a reasoning path that has terminated at the wrong answer. In our setting this is when the module emits the \texttt{/boxed\{incorrect answer\}}. This is an absorbing region. 
    \end{itemize}
    \item \textbf{Verification Predicate (\texttt{solved}):}  A predicate function $\mathcal{S}\to 0,1$ judging if the right answer is already derivable from the given partial reasoning state. Note that this is a simple formatting action, and is independent of the module $m$. 
    \begin{itemize}
        \item \texttt{solved(s) = 0} $\forall s \in \mathcal{F}$
\item \texttt{solved(s) = 1} $\forall s \in \mathcal{G}$

    \end{itemize}
    \item  \textbf{Maximum Reasoning Steps (N)}: We rollout the reasoning process up to $N$ steps. After $N$ steps of reasoning, we enforce a \emph{model-independent} termination rule where the state deterministically goes into $s' \in \mathcal{G}$ if \texttt{solved(s)} = 1 and into $s' \in \mathcal{F}$ if \texttt{solved(s)} = 0. For simplicity of notation, we assume the total trajectory length to be $N+1$.
    \item \textbf{Action Space $(A)$}: For a fixed meta-policy $\pi_{Rec}$ that recursively generates steps until termination (such as the one used by baseline CoT or the ARM-only variant), the meta policy executes a single action at any give state $s\in \mathcal{U}$: i.e., invokes a step-generator module $m$ to produce the next reasoning step. Thus, the action space is a singleton $\mathcal{A}=\{\texttt{generate\_step}\}$. For terminal states $\mathcal{G}\cup \mathcal{F}$, this is a \texttt{no-op}.  Hence, the choice of the module $m$ fully defines the transition dynamics of the MDP.
    \item  \textbf{Reward Function $(R)$: }The one-shot terminal reward is sparse:
\begin{equation*}
R(s\to s') =
\begin{cases}
1, & s'\in \mathcal{G},\\
0, & \text{otherwise.}
\end{cases}
\end{equation*}

    \item \textbf{Transition Dynamics (P)}: We denote the state transition probability $P(.|s,m)$ with the Markov assumption. This simplification is the core foundation for our MDP analysis.
\item \textbf{Value Function:}  For $n\in\{0,\dots,N\}$, let $V_n^m(s)$ denote the value with $n$ reasoning steps remaining before the formatting step. The Bellman recursion can be written as
\begin{equation*}
V_n^m(s) =
\begin{cases}
1, & s\in \mathcal{G},\\[4pt]
0, & s\in \mathcal{F},\\[4pt]
\ident\{\texttt{solved}(s)\} + \ident\{\neg\texttt{solved}(s)\}\, \mathbb{E}_{s'\sim P_m(\cdot\mid s)}[V_{n-1}^m(s')], & s\in\mathcal{U},\, n\ge 1,\\[4pt]
\ident\{\texttt{solved}(s)\}, & s\in\mathcal{U},\, n=0.
\end{cases}
\end{equation*}

\end{itemize}

Within this MDP framework, the ideal objective is to discover a module $m^*$, that maximizes the expected value from the initial state distribution $d_0(s)$

$$m^*=\underset{m\in\mathcal{M}}{\text{arg\, max}\,}\mathbb{E}_{s_0\sim d_0(s)}\left[V^m_N(s_0)\right]$$
This objective poses several major optimization challenges: 1) credit assignment problem over long sequence of steps and 2) unconstrained search space of code modules. 

\subsection{Definitions}
We introduce the following quantities to characterize module's performance and search strategy:
\begin{itemize}
    \item \textbf{Per-step competence $\delta_m(s)$:} This represents the \emph{competence} of the module $m$ at a reasoning state $s\in \mathcal{U}$ analogous to $\delta_{t,n}$ term in \cite{chang2025rlstartheoreticalanalysisreinforcement}. The probability that a one step update $s\in\mathcal{U}$ is valid can be viewed as a monotonically increasing function over $\delta_m(.)$, but for simplicity of notation, we assume this to be $\delta_m(s)$ itself. 
\item \textbf{Recovery $r_m(s)$:} The probability that, conditioned on an invalid one-step update from $s$, the next step returns to a valid state. This term captures the recovery possibility of a mistake in the immediate next turn. While recovery can happen at any turn following the mistake in a real LLM, we limit the window to 1 for simplicity. 
\item \textbf{Composite Validity $\phi_m(s)$:} The total probability that the next step is valid for a step $s\in\mathcal{U}$, either by being immediately valid, or by being successfully repaired on the next step:
\begin{equation}
    \phi_m(s)=\delta_m(s) + (1-\delta_m(s))r_m(s)
\end{equation}
\item \textbf{Window $W=(n,l)$:} A block of $l$ consecutive steps starting at step index $n$ where all states remain in $\mathcal{U}$. The ARM module replaces the baseline module $m_{CoT}$ with a candidate module $m$ only on this window. 
\item \textbf{Visitation Weights $w^\pi(W)$:} The probability under {baseline} policy $(\pi, m_{CoT})$ that the window $W$ occurs. This measures the frequency with which the meta-policy starts the module at a given window. 
\end{itemize}

\subsection{Key Simplifying Assumptions}
The rest of our analysis relies on the following key assumptions.
\begin{assumption}
\label{assumption_local_competence}
[Local competence lift in the scaffolding window]
Within any given window $W=(n,l)$, for all states $s$ visited, the candidate model satisfies $\phi_m(s) \geq \phi_{m_{CoT}(s)} + \Delta_c$ for some lift $\Delta_c\in [0,1)$.

\textit{Rationale:} This is the empirical premise that our scaffolded search objective (Section~\ref{method_step_gen}) is designed to optimize for. Our Algorithm~\ref{alg:rets}  directly measures and selects for modules that improve local validity and recovery rates over the CoT within a constrained context at random locations.
\end{assumption}

\begin{assumption}
\label{assumption_compatibility_loss}
[Compatibility Loss]
We define $\beta_l(W)$ a bound on the probability that replacing the baseline $m_{CoT}$ with ARM module $m$ at a window $W$ yields a context which is unusable for the rest of the baseline reasoning trace. We refer to $(1-\beta_l(W))$ as the \emph{compatibility factor}.  Furthermore, we define
$\overline{\beta}_l \coloneqq \sup_{W \in \mathcal{W}_{\text{valid}}} \beta_l(W)$
as the supremum of incompatibility probabilities across all valid windows, representing the worst-case incompatibility bound.

\textit{Rationale:} Swapping the baseline module $m_{CoT}$ with $m$ can introduce a ``\emph{context drift}'' or ``\emph{semantic drift}'' which could amplify at deployment time when the ARM module $m$ is used through the entire trajectory. Our approach minimizes this drift by two means: 1) the few shot examples of the progress, as well as the provided partial progress acts as a powerful inductive bias to constrain the next step to states that preserve usefulness, for example by adopting the same notation, logical continuity, etc. 2) the reviewer agent which proposes mutations to the module (starting from baseline CoT) is prompted to generate modules which solve one step at a time starting from the given partial progress.  
\end{assumption}

\begin{assumption}
    \label{assumption_termination}
    [Module-invariant termination]
    We assume that the reward is terminal, and is provided under the condition that the extracted answer matches the right answer. Furthermore, the last step in the MDP is reserved for this extraction, which is considered module-invariant, i.e. both CoT or any other module can do this final syntax step (equally) perfectly. 
    
\end{assumption}
\subsection{Theoretical Grounding for the Scaffolded Step-Generator Search}
\label{sec:apdx_scaffolding_step_generator_theory
}
The scaffolded objective evaluates a candidate $m$ by \emph{splicing} it into a baseline rollout for a short window
$t\in\{i,\dots,i+\ell-1\}$ while keeping $m_{\text{CoT}}$ before and after:
\[
\underbrace{U^{*}_{m_{\text{CoT}}}\circ \big(\,U^\ell_{m}\,\big)\circ U^{i}_{m_{\text{CoT}}}}_{\text{“baseline–candidate–baseline”}}.
\]
This section formalizes the link between local module improvements and global performance gains.
Under our simplified MDP framework and assumptions~\ref{assumption_local_competence}, \ref{assumption_compatibility_loss},  we establish the following lemmas:

\begin{lemma}
\label{lemma_window_lift}
[Window lift from local competence]
The probability of remaining in $\mathcal{U}$ after the window increases by at least $l.C^{l-1}.\Delta_c$ for some constant $C\in(0, 1)$. 

\textit{Proof:} The $l$ step window survival (being in $\mathcal{U}$) is at least $\phi_m(s)^l$. Under Assumption~\ref{assumption_local_competence}, the per-step composite validity improves by at least $\Delta_c$.  Hence, the survival probability improves by at least $\bigl(\phi_{\mathrm{CoT}}(s) + \Delta_c\bigr)^{l} - \phi_{\mathrm{CoT}}(s)^{l}$.  Applying the mean value theorem on $f(x)=x^\ell$, we get $$(x+\Delta_c)^\ell - x^\ell = \ell \xi^{\ell-1} \Delta_c \geq \ell (\min_{s\in \mathcal{U_R}} \phi_{\mathrm{CoT}}(s))^{\ell-1} \Delta$$ for some $\xi \in [x, x+\Delta]$, where $\mathcal{U}_R$ represents states reachable by baseline $CoT$ module. Further, let $C:=\min_{s\in U_R}\phi_{\mathrm{CoT}}(s)$ be a constant.


\end{lemma}
\begin{lemma}
    \label{lemma_compatibility}
    [Accounting for Compatibility]
    The probability of a sample surviving the window $W$ and remain usable after the module swap is lower bounded by $(1-\beta_l(W)).(l.C^{l-1}.\Delta_c)$ where C is the constant from Lemma~\ref{lemma_window_lift}. 

\textit{Proof:}  From Assumption~\ref{assumption_compatibility_loss}, the usability probability is at least $(1-\beta_l(W))$. Multiplying this by the probability of surviving the window from Lemma~\ref{lemma_window_lift} yields the result.
 \end{lemma}
\begin{lemma}
    \label{lemma_full_rollout}
    [From window survival to finite-horizon success]
    Any increase in the probability of staying within $\mathcal{U}$ across a window $W$ (while remaining usable) weakly increases the probability of reaching $\mathcal{G}$ within the horizon $N$.

    \textit{Proof:} Under assumption~\ref{assumption_termination}, the termination rule is module-invariant, and reaching the goal state only depends on being in state $s\in \mathcal{U}\cup\mathcal{G}$ and \texttt{solved(s)=1}. Thus a higher probability of preserving valid, in-progress states across a window cannot decrease (and generally increases) the likelihood that subsequent steps will generate such a solvable state before the horizon is exhausted. This follows from standard monotonicity arguments on absorbing Markov chains.

\end{lemma}
\begin{theorem}
    [Gain from Scaffolded Module Substitution in recursive meta policy]
    Let $J(\pi_{Rec}, m)=\E\left[V_N^{\pi_{Rec}, m}(s_0)\right]$ denote the expected terminal reward (success probability) obtained when recursively applying the step-generator module $m$ under a fixed baseline meta-policy $\pi_{Rec}$ for horizon $N$. The improvement of the ARM module $m^*$ over $m_{CoT}$ is at least: 
\begin{equation}
    J(\pi_{Rec},m^{*}) - J(\pi_{Rec},m_{\text{CoT}}) 
    \;\geq\; 
    \sum_{W} w_{\pi}(W) \, \kappa(W) \, (1 - \beta_{l}(W)) \, l \, C^{l-1} \, \Delta_{c}.
\end{equation}
where each term represents:
\begin{itemize}
    \item $w_\pi(W)$: visitation probability of a window $W$ under a baseline rollout;
    \item $\kappa(W)$: probability that a usable post-window state leads to terminal success within the remaining horizon. 
    \item $C\in(0,1]$: a constant from Lemma~\ref{lemma_window_lift} capturing compounding survival over steps.
\end{itemize}
In particular, if $\kappa(W)\geq \kappa_{min}\geq 0$ for all $W$, then
\begin{equation}
    J(\pi_{Rec},m^{*}) - J(\pi_{Rec},m_{\text{CoT}}) 
    \;\geq\; 
     \kappa_{min} \, (1 - \beta_{l}(W)) \, l \, C^{l-1} \, \sum_{W} w_{\pi}(W) \
\end{equation}
\textit{Proof:} From Lemma~\ref{lemma_compatibility}
, the probability of surviving the window is lower bounded by $(1-\beta_l)lC^{l-1}$. Let $\kappa(W)$ represent the probability of success upon starting from a good state, post the window. By Lemma~\ref{lemma_full_rollout}, the increase in usable post-window mass  translates to atleast a $\kappa(W)$ fraction improvement in terminal success within the remaining horizon. Thus the expected gain from the window is $\kappa(W)(1-\beta_l(W))lC^{l-1}$. Taking expectation over window visitation probabilities yields the result:
\begin{equation}
\begin{aligned}
    J(\pi_{Rec},m^{*}) - J(\pi_{Rec},m_{\text{CoT}}) 
    &= \sum_{W} w_{\pi}(W) \cdot \text{Gain}(W) \\
    &= \sum_{W} w_{\pi}(W) \cdot \kappa(W)\,(1 - \beta_{l}(W))\, l\, C^{l-1}. \\
    &\geq \sum_{W} w_{\pi}(W) \cdot \kappa_{min}\,(1 - \beta_{l}(W))\, l\, C^{l-1}.
\end{aligned}
\end{equation} 

where $k_{min}\geq 0$ is the lowest probability of success from a valid, usable intermediate reasoning trace. 
\end{theorem}

\subsection{Theoretical Justification for zero-shot policy transfer}
The learned meta policy $\pi^*$ uses  $m_{CoT}$ as the \emph{step generator} during the learning phase and is deployed zero-shot using the discovered ARM module $m^*$. Below, we justify why this transfer is effective. 

\begin{theorem}
    [Validity of Zero-shot step generator swap in Meta policy]
     Let $J(\pi^*, m^*)=\E_{s_0\sim \mathcal{D}}\left[V_N^{(\pi^*, m^*)}(s_0)\right]$ denote the expected terminal reward obtained when applying the discovered meta-policy $\pi^*$ (from Section~\ref{method_meta_policy}), with the step-generator module $m^*$ under horizon $N$. If $\Delta_c\geq \frac{\overline{\beta_l}}{1-\overline{\beta_l}}$, then the transfer is valid, i.e., $J(\pi^*, m^*)\geq J(\pi^*, m_{CoT})$
\end{theorem}
Let's define per-step advantage of a module $m$ over $m_{CoT}$ with $n$ more steps to go as the expected difference in value when taking one step with $m$ and the rest with $m_{CoT}$:
\begin{equation}
    A_n(s_n, m)\triangleq \E_{s'\sim P_m(.|s)}[V^{(\pi^*, m_{CoT})}_{n-1}(s')] - \E_{s'\sim P_{m_{CoT}}(.|s)}[V^{(\pi^*, m_{CoT})}_{n-1}(s')]
\end{equation}
Now let's consider the difference in expected value starting from a given state $s_0$ sampled from the data distribution $\mathcal{D}$. For simplicity, we drop $\pi^*$ from notation as it is the common meta policy in both terms. 
$$V_N^m(s_0) - V_N^{m_{CoT}}(s_0)$$
Rolling out for one step yields
$$\E_{s_1\sim P_m(.|s_0)}[V^ m_{n-1}(s_1)] - V_n^{m_{CoT}}(s_0)$$
Adding an subtracting $\E_{s_1\sim P_m(.|s_0)}[V^ {m_{CoT}}_{n-1}(s_1)]$ (i.e., sampling from $m$ but continue with $m_{CoT}$) we get:
\begin{equation}
\begin{aligned}
\E_{s_1\sim P_m(\cdot|s_0)}[V^{m_{CoT}}_{n-1}(s_1)]
&- \E_{s_1\sim P_{m_{CoT}}(\cdot|s_0)}[V^{m_{CoT}}_{n-1}(s_1)] \\
&+ \E_{s_1\sim P(\cdot|s_0)}[V_{n-1}^m(s_1) - V_{n-1}^{m_{CoT}}(s_1)]
\end{aligned}
\end{equation}

By Equation-5, this can be written as
$$A_n^{m_{CoT}}(s_0, m) + \E_{s_1\sim P(.|s_0)}[V_{n-1}^m(s_1)-V_{n-1}^{m_{CoT}}(s_1)]$$
This is a recursive equation in $n$ since the second term is the difference in value between the module with $n-1$ steps to horizon. Hence:
\begin{equation}
\begin{aligned}
    V_N^m(s_0) - V_N^{m_{CoT}}(s_0) &= \sum_{n=0}^{N}\bigl[ A_{N-n}^{m_{CoT}}(s_n, m)\bigr]
\end{aligned}
\end{equation}
Thus we can conservatively guarantee module improvement, if each of the the advantage term is positive. Suppose that $U$ represents the event that one step rollout using our discovered module $m^*$ is usable (i.e. no errors, and usable context) in the next turn, then by law of total expectation the advantage term can be written as:

\begin{align}
&\mathbb{P}(U \mid s, m^*) 
  \cdot \big(
    \E[V_{N-n} \mid s, m^*, U] 
    - \E[V_{N-n} \mid s, m_{\text{CoT}}, U]
  \big) \nonumber \\
&\quad +\; 
  \mathbb{P}(\neg U \mid s, m^*) 
  \cdot \big(
    \E[V_{N-n} \mid s, m^*, \neg U] 
    - \E[V_{N-n} \mid s, m_{\text{CoT}}, \neg U]
  \big)
\end{align}

By Assumption~\ref{assumption_local_competence} and Assumption~\ref{assumption_compatibility_loss}, the first term is at least $(1-\overline{\beta_l})\cdot \Delta_c$. The second term is lower bounded in the worst case by $\overline{\beta_l}\cdot (-1)$ since the probability of non-useful state is $\overline{\beta_l}$ and the difference in reward is at most $-1$ (when $m^*$ fails where $m_{CoT}$ succeeds). Thus a conservative lower bound is:
\begin{equation}
    \begin{aligned}
    A_n^{m_{CoT}}(s, m) &\geq     (1-\overline{\beta_l})\cdot \Delta_c - \overline{\beta_l}
    \end{aligned}
\end{equation}
Hence advantage is non-negative when $\Delta_c\geq \frac{\overline{\beta_l}}{1-\overline{\beta_l}}$, and we can guarantee that zero-shot transfer is effective when this condition is satisfied. 

\begin{corrolary}
[] The meta policy, paired with ARM, $(\pi^*, m^*)$ yields the highest performance when $\Delta_c$ is large and $\beta_l$ is small. 

\emph{Remark:} From Equation-8, the claim holds. Our scaffolded objective function is precisely designed to maximize $\Delta_c$ while reducing the probability of state drift leading to mismatched compatibility. In \ref{fig:arm_analyses}, we can empirically see that better performance is associated with improvement in per-step competence. And in \ref{tab:meta_analysis}, we see that the ARM module is compatible with $m_{CoT}$ and can act as a superior drop-in replacement for CoT. Furthermore, the examples shown in Section~\ref{sec:step_rollout_examples} show that both CoT and ARM rollouts are  empirically non-discriminable. 
\end{corrolary}

\tcbset{
  myblock/.style={
    colback=gray!5!white,
    colframe=gray!50!black,
    boxrule=0.5pt,
    left=2mm, right=2mm, top=1mm, bottom=1mm,
    before skip=0pt, after skip=0pt,
    sharp corners,
  }
}

\section{Step Generator roll out Examples}
\label{sec:step_rollout_examples}
\subsection{Example-1}
\begin{tcolorbox}[myblock, title={ARM Rollout}]
\begin{lstlisting}[language=, caption=, basicstyle=\ttfamily\small, breaklines=true]
### Step 2
Express the sides in terms of the common ratio \(2R\):
\[
a = 2R \sin 60^\circ = 2R \times \frac{\sqrt{3}}{2} = R \sqrt{3},
\]
\[
b = 2R \sin 45^\circ = 2R \times \frac{\sqrt{2}}{2} = R \sqrt{2},
\]
\[
c = 2R \sin 75^\circ.
\]
Recall that \(\sin 75^\circ = \sin(45^\circ + 30^\circ)\):
\[
\sin 75^\circ = \sin 45^\circ \cos 30^\circ + \cos 45^\circ \sin 30^\circ = \frac{\sqrt{2}}{2} \times \frac{\sqrt{3}}{2} + \frac{\sqrt{2}}{2} \times \frac{1}{2} = \frac{\sqrt{6}}{4} + \frac{\sqrt{2}}{4} = \frac{\sqrt{6} + \sqrt{2}}{4}.
\]
Hence,
\[
c = 2R \times \frac{\sqrt{6} + \sqrt{2}}{4} = R \times \frac{\sqrt{6} + \sqrt{2}}{2}.
\]
\end{lstlisting}
\end{tcolorbox}

\begin{tcolorbox}[myblock, title={CoT Rollout}]
\begin{lstlisting}[language=, caption=, basicstyle=\ttfamily\small, breaklines=true]
### Step 2
Expressing the sides in terms of the circumradius \( R \):
\[
a = 2R \sin A = 2R \sin 60^\circ = 2R \times \frac{\sqrt{3}}{2} = R \sqrt{3},
\]
\[
b = 2R \sin B = 2R \times \frac{\sqrt{2}}{2} = R \sqrt{2},
\]
\[
c = 2R \sin C = 2R \times \sin 75^\circ.
\]
Recall that:
\[
\sin 75^\circ = \sin (45^\circ + 30^\circ) = \sin 45^\circ \cos 30^\circ + \cos 45^\circ \sin 30^\circ = \frac{\sqrt{2}}{2} \times \frac{\sqrt{3}}{2} + \frac{\sqrt{2}}{2} \times \frac{1}{2} = \frac{\sqrt{6}}{4} + \frac{\sqrt{2}}{4} = \frac{\sqrt{6} + \sqrt{2}}{4}.
\]
Therefore:
\[
c = 2R \times \frac{\sqrt{6} + \sqrt{2}}{4} = R \times \frac{\sqrt{6} + \sqrt{2}}{2}.
\]
\end{lstlisting}
\end{tcolorbox}

\subsection{Example-2}
\vspace{-2.0em} 
\noindent

\begin{tcolorbox}[myblock, title={ARM Rollout}]
\begin{lstlisting}[language=, caption=, basicstyle=\ttfamily\small, breaklines=true]
### Step 3
Now, determine the new temperature after doubling the kinetic energy:
\[
T_2 = 526.3\,K
\]
Convert this back to Celsius:
\[
T_2^{\circ} C = 526.3 - 273.15 = 253.15^\circ C
\]
This is a significant increase from the initial \(-10^\circ C\), indicating the molecules now have enough energy corresponding to about \(253^\circ C\).
\end{lstlisting}
\end{tcolorbox}
\begin{tcolorbox}[myblock, title={CoT Rollout}]
\begin{lstlisting}[language=, caption=, basicstyle=\ttfamily\small, breaklines=true]
### Step 3
Determine the new temperature after doubling the kinetic energy:
Since kinetic energy \( \propto T \),
\[
T_2 = 2 \times T_1 = 526.3 \, K
\]
Convert this back to Celsius:
\[
T_2 (^\circ C) = T_2 (K) - 273.15 = 526.3 - 273.15 \approx 253.15^\circ C
\]
This temperature is well above the boiling point of water at 1 atm (100C). Therefore, the molecules' kinetic energy would be sufficient to overcome intermolecular forces and convert the water from the solid phase (ice) into the gas phase (vapor).
\end{lstlisting}
\end{tcolorbox}

\subsection{Example-3}

\begin{tcolorbox}[myblock, title={ARM Rollout}]
\begin{lstlisting}[language=, caption=, basicstyle=\ttfamily\small, breaklines=true]
### Step 1

To determine the correct substitution from equation 12.9 to equation 12.10, we need to understand the context of these equations and the variables involved. Equation 12.9 likely involves a momentum-like variable \( P^+ \) or a gamma matrix that acts on spinors, and the goal is to simplify or rewrite this expression into a form involving gamma matrices or energy-momentum components.

Typically, in relativistic quantum mechanics or quantum field theory, the transformation from the light-cone momentum \( P^+ \) to gamma matrices involves choosing a specific gamma matrix to project or decompose spinor components. The options provided suggest that the substitution involves gamma matrices and possibly the momentum operator or gamma matrices related to time.

Given that, the known structures are:

- \( P^+ \): light-cone momentum component, usually \( P^+ = P^0 + P^3 \).
- \( \gamma^u \): a gamma matrix with a specific index \( u \), possibly a spatial index.
- \( \gamma^0 \): the gamma matrix corresponding to time.
- \( \gamma^0 P \): a product of \( \gamma^0 \) with some momentum operator \( P \).

Since the transformation is from equation 12.9 to 12.10, and the options involve gamma matrices and momenta, the typical approach in such derivations involves rewriting the projection operator or the momentum in terms of gamma matrices.
\end{lstlisting}
\end{tcolorbox}

\begin{tcolorbox}[myblock, title={CoT}]
\begin{lstlisting}[language=, caption=, basicstyle=\ttfamily\small, breaklines=true]
### Step 1

To determine the correct substitution, first understand the context: equation 12.9 involves the momentum operator \( P^\mu \), and equation 12.10 is likely a transformed version involving gamma matrices. In relativistic quantum mechanics, transformations often involve gamma matrices \( \gamma^\mu \) and the gamma matrices' role in formulating Dirac equations.

Given the options:

- A: \( P^+ \)
- B: \( \gamma^u \)
- C: \( \gamma^0 \)
- D: \( \gamma^0 P \)

and based on the typical transformations, the substitution from \( P^\mu \) to a form involving gamma matrices usually involves \( \gamma^0 \) because it relates to the time component of the gamma matrices and appears frequently in the Dirac formalism.

In particular, if the original equation involves \( P^\mu \), and the transformed equation involves a gamma matrix, then the key is recognizing the common substitution that converts momentum operators into gamma matrix expressions, especially the time component \( P^0 \).

Therefore, the step is to analyze which gamma matrix substitution correctly transforms the operator form in equation 12.9 into the one in 12.10.
\end{lstlisting}
\end{tcolorbox}
\newpage
\section{Best ARM discovered: CriticChainOfThoughtV7}
\label{apdx:discovered_arms}

The following is the Python implementation of the best ARM discovered by our algorithm.


\begin{lstlisting}[style=pywrap, caption={Code for CriticChainOfThoughtV7, performance: 38.0}]
import asyncio

class CriticChainOfThoughtV7:
    def __init__(self, llm):
        self.llm = llm

    async def forward(self, problem, partial_progress):
        # 1. Generate four candidate next steps in parallel
        candidate_tasks = [
            self.llm.generate_step(problem, partial_progress)
            for _ in range(4)
        ]
        candidates = await asyncio.gather(*candidate_tasks)

        # 2. Critique candidates in two groups of two, in parallel
        critique_tasks = []
        groups = [
            (0, 2, ("rating_1", "rating_2"), ("critique_1", "critique_2")),
            (2, 4, ("rating_3", "rating_4"), ("critique_3", "critique_4"))
        ]
        for start, end, rating_names, critique_names in groups:
            context = [
                {
                    "name": "Problem",
                    "data": problem,
                    "description": "The problem to solve."
                },
                {
                    "name": "Partial Progress",
                    "data": partial_progress,
                    "description": "The partial solution so far."
                },
                {
                    "name": "Candidate Next Steps",
                    "data": "\n\n".join(
                        f"### Candidate Next Step {i+1}\n{candidates[i]}"
                        for i in range(start, end)
                    ),
                    "description": "Two candidate next steps formatted with markdown subheaders."
                }
            ]
            instructions = (
                "You are given a problem, the current partial solution, and two candidate next reasoning steps.\n"
                "For each candidate, provide:\n"
                f"- {rating_names[0]} and {rating_names[1]}: a single integer rating from 1 to 10 indicating its fit as the next reasoning step (10 is best).\n"
                f"- {critique_names[0]} and {critique_names[1]}: a one-sentence critique highlighting each candidate's strengths and weaknesses.\n"
                f"Name the fields exactly {rating_names[0]}, {critique_names[0]}, {rating_names[1]}, {critique_names[1]}."
            )
            response_format = [
                {
                    "name": rating_names[0],
                    "description": f"Integer rating (1-10) for Candidate Next Step {start+1}."
                },
                {
                    "name": critique_names[0],
                    "description": f"One-sentence critique of Candidate Next Step {start+1}."
                },
                {
                    "name": rating_names[1],
                    "description": f"Integer rating (1-10) for Candidate Next Step {start+2}."
                },
                {
                    "name": critique_names[1],
                    "description": f"One-sentence critique of Candidate Next Step {start+2}."
                }
            ]
            critique_tasks.append(
                self.llm.chat_completion(context, instructions, response_format)
            )

        critiques = await asyncio.gather(*critique_tasks)

        # 3. Parse ratings and identify the two highest-rated candidates
        ratings = [
            int(critiques[0]["rating_1"]),
            int(critiques[0]["rating_2"]),
            int(critiques[1]["rating_3"]),
            int(critiques[1]["rating_4"])
        ]
        sorted_indices = sorted(range(4), key=lambda i: ratings[i], reverse=True)
        top1_idx, top2_idx = sorted_indices[0], sorted_indices[1]
        top1_candidate = candidates[top1_idx]
        top2_candidate = candidates[top2_idx]

        # 4. Final head-to-head comparison between the top two candidates
        context_final = [
            {
                "name": "Problem",
                "data": problem,
                "description": "The problem to solve."
            },
            {
                "name": "Partial Progress",
                "data": partial_progress,
                "description": "The partial solution so far."
            },
            {
                "name": "Candidate Next Steps",
                "data": (
                    f"### Candidate A\n{top1_candidate}\n\n"
                    f"### Candidate B\n{top2_candidate}"
                ),
                "description": "Two top candidate next steps formatted with markdown subheaders."
            }
        ]
        instructions_final = (
            "Compare Candidate A and Candidate B as the next reasoning step for the given problem and partial progress.\n"
            "Provide:\n"
            "- winner: choose either 'Candidate A' or 'Candidate B' indicating which step is better.\n"
            "- justification: one-sentence justification for your choice."
        )
        response_format_final = [
            {
                "name": "winner",
                "description": "Either 'Candidate A' or 'Candidate B' indicating the better next step."
            },
            {
                "name": "justification",
                "description": "One-sentence justification for the choice."
            }
        ]
        final_decision = await self.llm.chat_completion(
            context_final, instructions_final, response_format_final
        )

        if final_decision["winner"].strip() == "Candidate A":
            selected_candidate = top1_candidate
            runnerup_candidate = top2_candidate
        else:
            selected_candidate = top2_candidate
            runnerup_candidate = top1_candidate

        # 5. Post-selection adversarial critique with severity rating
        context_flaw = [
            {
                "name": "Problem",
                "data": problem,
                "description": "The problem to solve."
            },
            {
                "name": "Partial Progress",
                "data": partial_progress,
                "description": "The partial solution so far."
            },
            {
                "name": "Selected Candidate Next Step",
                "data": f"### Selected Candidate Next Step\n{selected_candidate}",
                "description": "The final chosen candidate next reasoning step formatted with a markdown subheader."
            }
        ]
        instructions_flaw = (
            "You are given a problem, the current partial solution, and a selected next reasoning step.\n"
            "Identify any major flaw or missing piece of reasoning in the selected step.\n"
            "Provide:\n"
            "- flaw: either the single word 'None' if there is no flaw, or a brief description of the flaw.\n"
            "- severity: a single integer rating from 1 to 10 indicating how severe the flaw is (10 is critical)."
        )
        response_format_flaw = [
            {
                "name": "flaw",
                "description": "Either the single word 'None' if there is no flaw, or a brief description of a major flaw in the selected step."
            },
            {
                "name": "severity",
                "description": "Integer rating (1-10) indicating severity of the flaw (10 is most severe)."
            }
        ]
        flaw_response = await self.llm.chat_completion(
            context_flaw, instructions_flaw, response_format_flaw
        )
        flaw = flaw_response["flaw"].strip()
        severity = int(flaw_response["severity"])

        # 6. Compute dynamic severity threshold based on rating gap
        gap = ratings[top1_idx] - ratings[top2_idx]
        if gap <= 1:
            threshold = 5
        elif gap == 2:
            threshold = 6
        else:
            threshold = 7

        # 7. If a severe flaw is detected above the dynamic threshold, fall back
        if flaw.lower() != "none" and severity >= threshold:
            return runnerup_candidate
        return selected_candidate
\end{lstlisting}

\section{Best Meta-Policy Discovered: VerifiedWeightedAdaptiveSelfConsistentChainOfThought}
\label{apdx:discovered_metapolicies}

The following is the Python implementation of the best meta-policy discovered by our algorithm.

\begin{lstlisting}[style=pywrap, caption={Code for VerifiedWeightedAdaptiveSelfConsistentChainOfThought, performance: 38.0}]
import asyncio
from agent.solution import Solution, Step
from judge_utils import judge_equality

class VerifiedWeightedAdaptiveSelfConsistentChainOfThought:
    def __init__(self, llm, block):
        self.llm = llm
        self.block = block

    async def forward(self, problem):
        # Helper: generate one chain up to 8 steps, then complete via LLM if needed
        async def generate_chain():
            solution = Solution()
            for _ in range(8):
                next_step = await self.block.forward(problem, str(solution))
                solution.add_step(Step(str(next_step)))
                if solution.is_completed():
                    return solution
            completion = await self.llm.complete_solution(problem, str(solution))
            solution.add_step(Step(str(completion)))
            return solution

        # Helper: confidence scoring (1-5)
        async def score_chain(chain):
            context = [
                {"name": "Problem", "data": problem, "description": "The original problem statement."},
                {"name": "Chain",   "data": str(chain), "description": "Full chain-of-thought reasoning plus final answer."}
            ]
            instructions = (
                "You are evaluating the chain-of-thought solution for the given problem. "
                "On a scale from 1 (very uncertain) to 5 (very confident), rate your confidence "
                "that the final answer is correct. Output ONLY the integer confidence (1-5)."
            )
            response_format = [{"name": "Confidence", "description": "Integer from 1 to 5"}]
            resp = await self.llm.chat_completion(context, instructions, response_format)
            # parse safely
            try:
                conf = int(resp["Confidence"].strip())
            except Exception:
                conf = 1
            return max(1, min(conf, 5))

        # Helper: verify logical consistency (Yes/No)
        async def verify_chain(chain):
            context = [
                {"name": "Problem", "data": problem, "description": "The original problem statement."},
                {"name": "Chain",   "data": str(chain), "description": "Full chain-of-thought reasoning plus final answer."}
            ]
            instructions = (
                "Review the chain-of-thought reasoning for the given problem. "
                "Is the reasoning free of logical errors or contradictions? "
                "Output ONLY 'Yes' if it is fully logical, otherwise output 'No'."
            )
            response_format = [{"name": "Valid", "description": "Yes or No"}]
            resp = await self.llm.chat_completion(context, instructions, response_format)
            valid = resp.get("Valid", "").strip().lower().startswith("y")
            return valid

        # Weighted vote helper
        def find_best_weighted(chains_list, conf_list):
            weight_sums = {}
            total = sum(conf_list)
            for chain, cf in zip(chains_list, conf_list):
                ans = chain.answer()
                weight_sums[ans] = weight_sums.get(ans, 0) + cf
            best_ans, best_w = None, -1
            for ans, w in weight_sums.items():
                if w > best_w:
                    best_ans, best_w = ans, w
            return best_ans, best_w, total

        # 1) Generate initial 3 chains in parallel
        initial = [generate_chain() for _ in range(3)]
        chains = await asyncio.gather(*initial)

        # 2) Score and verify each chain
        score_tasks = [score_chain(ch) for ch in chains]
        verify_tasks = [verify_chain(ch) for ch in chains]
        confidences = await asyncio.gather(*score_tasks)
        valids = await asyncio.gather(*verify_tasks)

        max_chains = 7

        # 3) Adaptive sampling with verification gating
        while True:
            # Determine which chains to consider: only verified if any, else all
            if any(valids):
                considered_chains = [ch for ch, v in zip(chains, valids) if v]
                considered_confs   = [cf for cf, v in zip(confidences, valids) if v]
            else:
                considered_chains = chains
                considered_confs   = confidences

            best_ans, best_weight, total_weight = find_best_weighted(considered_chains, considered_confs)
            # stop if weighted majority reached or chain cap
            if best_weight > total_weight / 2 or len(chains) >= max_chains:
                break

            # else generate one more chain, score & verify, then loop
            new_chain = await generate_chain()
            chains.append(new_chain)
            new_conf = await score_chain(new_chain)
            confidences.append(new_conf)
            new_valid = await verify_chain(new_chain)
            valids.append(new_valid)

        # 4) Select final chain: consensus & highest confidence among considered
        if any(valids):
            final_pool = [ (ch, cf) for ch, cf, v in zip(chains, confidences, valids) if v and judge_equality(ch.answer(), best_ans) ]
        else:
            final_pool = [ (ch, cf) for ch, cf in zip(chains, confidences) if judge_equality(ch.answer(), best_ans) ]

        selected_chain = None
        top_conf = -1
        for ch, cf in final_pool:
            if cf > top_conf:
                selected_chain, top_conf = ch, cf

        # Fallback if nothing selected
        if selected_chain is None:
            selected_chain = chains[-1]

        return selected_chain
\end{lstlisting}
\section{Cost Analysis}

\subsection{Training Cost}

Table \ref{tab:training_cost} shows the total training cost per iteration for each of the automated MAS design methods benchmarked. Note that while the ARM training cost is ~3.8x that of ADAS and ~3.9x that of AFlow

\begin{table*}[!ht]
\centering
\small
\resizebox{0.3\textwidth}{!}{%
\begin{tabular}{l c}
\toprule
\textbf{Model} & \textbf{Cost (USD)} \\ 
\midrule
ARM (Ours) & 4.53 \\
Meta (Ours) & 5.40 \\
Total Cost (Ours) & 9.93 \\
\midrule
ADAS & 2.62 \\
AFlow & 2.52 \\
\bottomrule
\end{tabular}
}
\vspace{2mm}
\caption{Training costs in USD of automated MAS design methods, when optimized on the 1000-sample validation dataset used for main results in Table \ref{tab:main_results}.}
\label{tab:training_cost}
\end{table*}

\subsection{Inference Cost}

Table \ref{tab:inference_cost} shows the inference cost for each of the automated MAS design methods benchmarked, for each evaluation dataset.

\begin{table*}[!ht]
\centering
\small
\resizebox{0.5\textwidth}{!}{%
\begin{tabular}{l c c c c}
\toprule
\textbf{Method} & \textbf{AIME} & \textbf{HMMT} & \textbf{GPQA} & \textbf{LiveBench} \\ 
\midrule
ARM (Ours) & 3.69 & 3.02 & 0.81 & 1.12 \\
ARM + Meta & 17.60 & 13.77 & 3.22 & 4.36 \\
\midrule
ADAS & 0.87 & 0.88 & 0.19 & 0.28 \\
AFlow & 0.78 & 0.66 & 0.16 & 0.19 \\
\bottomrule
\end{tabular}
}
\vspace{2mm}
\caption{Inference costs in USD of automated MAS design methods for each evaluation dataset. GPT-4.1-nano was used as the LLM for these experiments.}
\label{tab:inference_cost}
\end{table*}
\section{Reproducibility Statement}
\label{sec:reproducibility}

Upon publication, we commit to releasing the open-source code for our framework, including all discovered Agentic Reasoning Modules, meta-policies, and the specific prompts used for the Reviewer Agent. Our experiments were conducted using a mix of closed and open-source models. The MAS designer utilized OpenAI's \texttt{o4-mini-high} The reasoning modules were executed on \
\texttt{GPT-4.1-nano}, \texttt{GPT-4o}, and the open-source \texttt{Llama-3.3-70B}. All evaluation benchmarks, including MATH500, AIME, and HMMT, are publicly available.

\subsection{ARM Implementation Details}

The 1000-sample subset of Open-R1-Mixture-of-Thoughts was created by taking the math and science splits of the original dataset, filtering to samples which the provided Deepseek-R1 reasoning trace had length between 8k to 10k tokens (to filter to samples of appropriate difficulty), and randomly sampling 1000 problems from the filtered problems.

We run both the step-generator module optimization and the meta-policy optimization for 20 iterations. Both optimizations are performed using GPT-4.1-nano as the MAS executor model.

Whenever sampling from the MAS executor model, we use a temperature of 1.0 with a top\_p of 0.95.

\subsection{Baseline Implementation Details}

As in the ARM implementation, whenever sampling from the MAS executor model, we use a temperature of 1.0 with a top\_p of 0.95.

\begin{itemize}
    \item CoT: We use a simple CoT prompt that instructs the model to reason step-by-step and follow the final answer format.
    \item CoT-SC: We use $n=12$ parallel reasoning traces.
    \item Self-Refine: We limit to a maximum of 5 self refining iterations.
    \item LLM-Debate: We use 4 LLM agents debating for a maximum of 3 rounds.
    \item ADAS: We use the provided codebase, following the recommended run configuration. For a fair comparison to other baselines, we make a one line addition to the optimizer prompt to disallow arbitrary Python code execution within the discovered MASes, since other baselines do not utilize code execution. For the 1000-sample optimization, we use GPT-4.1-nano as the MAS executor model during optimization, following ARM's implementation.
    \item AFlow: We use the provided codebase, following the recommended run configuration. We allow the optimizer to utilize the Custom, AnswerGenerate, and ScEnsemble operators. For the 1000-sample optimization, we use GPT-4.1-nano as the MAS executor model during optimization, following ARM's implementation.
\end{itemize}


\end{document}